\documentclass{article}

\usepackage[preprint]{corl_2026} 
\usepackage{graphicx}
\usepackage{subcaption}
\usepackage{enumitem}
\usepackage{algorithm}
\usepackage{algpseudocode}
\usepackage{needspace}
\usepackage{amsmath}
 \usepackage{amssymb}
\usepackage{array}
\usepackage{booktabs}
\usepackage{multirow}
\graphicspath{{figures/}}

\title{EvoNav-Bench: Benchmarking Lifelong Navigation in Evolving Environments}

\author{%
\begin{minipage}{\dimexpr\textwidth-2\tabcolsep\relax}
\centering
{\small\bfseries
\mbox{Xilin Wang$^{1*}$,\enspace Guoxi Zhang$^{1*}$,\enspace Hongming Xu$^{1}$,\enspace Zhuofan Zhang$^{1,2}$,\enspace Tianxu Wang$^{1}$,\enspace Lifeng Fan$^{1\dagger}$}\endgraf}
\vspace{4pt}
{\footnotesize\normalfont
$^1$State Key Laboratory of General Artificial Intelligence, BIGAI\quad
$^2$Tsinghua University\endgraf}
\end{minipage}%
}

\begin{document}
\maketitle
\begingroup
\renewcommand{\thefootnote}{\fnsymbol{footnote}}
\footnotetext[1]{Equal contribution.\quad $^\dagger$Corresponding author.}
\endgroup


\begin{abstract}
Lifelong navigation (LN) requires an embodied agent to solve a sequence of navigation subtasks in the same environment.
Since solving each subtask from scratch incurs redundant exploration, an LN agent must consolidate experience from earlier stages and reuse it in later stages, often through persistent scene representations such as scene graphs or visual snapshots.
However, existing approaches typically assume a stationary environment, whereas in real-world LN settings, human activities can cause the environment to evolve.
With the stationary assumption violated, existing methods may fuse outdated prior observations with new observations, yet current benchmarks cannot reveal this failure mode.
In this paper, we present \textbf{EvoNav-Bench}, which extends the GOAT-Bench~\cite{Khanna_2024_CVPR} style LN formulation in the context of evolving environments.
Built on the ProcTHOR framework, EvoNav-Bench introduces environment modifications between navigation tasks, making prior experience useful but not fully reliable.
This design enables controlled evaluation of how environment evolution affects LN agents that reuse prior scene observations.
Using EvoNav-Bench, we benchmark three recent methods that build and reuse scene representations for navigation.
We also compare three simple heuristic strategies for handling environment evolution: Frontier-Update, Fail-then-Update, and Stage-Reset.
Our results show that existing methods are brittle under environment evolution, while the heuristic strategies enable a controlled analysis of how agents can adapt to scene changes and mitigate their impact.
\end{abstract}



\section{Introduction}
\label{sec:introduction}
\begin{figure}
    \centering
    \includegraphics[width=1.0\textwidth]{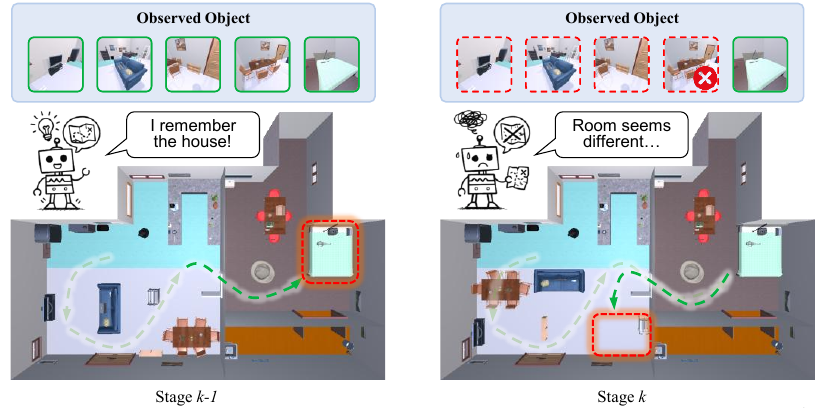}
    \caption{An illustration of lifelong navigation in an evolving environment. While navigating to the target object in stage $k-1$, the agent observes objects such as a TV, a sofa, a desk, and chairs, consolidating experience that can support future navigation. While the agent is in the bedroom, humans rearrange the living room. In this case, the agent should not blindly fuse new observations of the living room with prior experience. In such settings, prior observations can reduce redundant exploration, but they can also mislead later navigation when parts of the scene have evolved.}
    \label{fig:teaser}
    \vspace{-1.5em}
\end{figure}

Following GOAT-Bench~\cite{Khanna_2024_CVPR}, we consider lifelong navigation (LN)~\cite{Churchill_2012_ICRA} as a sequence of consecutive navigation tasks in which an agent learns about its environment and consolidates collected experiences.
The lifelong aspect highlights the importance of consolidating experience, since solving each stage from scratch causes redundant exploration.
This task-level formulation complements other views of lifelong navigation centered on continual policy learning~\cite{9345478}, cross-session adaptation~\cite{shi2019openlorisscene}, or persistent localization~\cite{1308004}.
Real-world systems such as GOAT~\cite{Chang_RSS_24} show that navigation performance improves as experience accumulates, suggesting the practical value of LN for long-term home robot deployment.

Existing LN methods often aggregate past observations into persistent scene representations, such as multi-view images~\cite{Yang_2025_CVPR}, feature maps~\cite{Ziliotto_2025_CVPR}, scene graphs~\cite{huang2026msgnavunleashingpowermultimodal}, or retrieved long-term experiences~\cite{wang2026trajragretrievinggeometricsemanticexperience,11459135}.
A common assumption behind these approaches is that the environment remains unchanged across stages.
For instance, 3D-Mem~\cite{Yang_2025_CVPR} directly fuses point clouds from new observations with existing ones when updating its scene representation.
\textit{However, lifelong navigation is distinguished by extended operation over time, during which the agent's visited environment may evolve due to human activity.}
Consider the assistant-robot scenario in Fig.~\ref{fig:teaser}, where a previously visited room is rearranged before a later navigation stage.
In this case, prior scene information can become outdated, while re-exploration from scratch remains inefficient because changes are often local.
This setting exposes a failure mode that remains largely hidden in static benchmarks: prior scene information can support efficient lifelong navigation, but it can also become misleading under environment evolution.
Environmental change itself has long been studied in dynamic and lifelong SLAMs~\cite{1308004,9345478,shi2019openlorisscene}; our focus is its interaction with cross-stage scene-representation reuse in goal-directed navigation.
Existing consecutive-navigation benchmarks~\cite{Khanna_2024_CVPR, NEURIPS2025_b78fef5a} still assume static environments, and thus do not systematically evaluate LN agents in evolving environments.

In this paper, we present \textbf{EvoNav-Bench}, which extends the GOAT-Bench style LN formulation with controlled environment evolution between navigation stages.
EvoNav-Bench is built on ProcTHOR~\cite{NEURIPS2022_27c546ab}, which offers a collection of procedurally generated indoor household environments.
Each episode in EvoNav-Bench takes place in a single unseen house environment and consists of multiple object navigation stages.
Between the stages, the environment is modified by relocating objects to plausible new positions in a room while preserving overall navigability.
Agents are not informed about environment evolution, so they need to detect and adapt to evolved scene states in subsequent tasks. 
Since evolution is applied probabilistically, EvoNav-Bench contains both static and evolving stage transitions.
In this way, EvoNav-Bench creates a LN setting where prior experience is useful for navigation but can be unreliable.

Using EvoNav-Bench, we benchmark three recent navigation agents, UniGoal~\cite{Yin_2025_CVPR}, 3D-Mem~\cite{Yang_2025_CVPR} and MSGNav~\cite{huang2026msgnavunleashingpowermultimodal} and find that they are brittle when the environment evolves.
We further compare three simple heuristic strategies for handling environment evolution: Frontier-Update, Fail-then-Update, and Stage-Reset.
The results suggest that re-exploration and revising storing object entries are effective mechanisms for adapting to scene evolution. Analysis on target relocation suggests the trade-off between reusing and reseting prior scene information.
A controlled analysis further analyzes how environment evolution can substantially affect LN agents that reuse accumulated scene information across stages, especially when retained information becomes outdated.
Experiments also show that perception quality can outweigh the influence of environment evolution.
In summary, our contributions are as follows. 
\begin{itemize}[leftmargin=1em]
    \item We extend the GOAT-style LN formulation with controlled between-stage environment evolution, enabling evaluation of its interaction with cross-stage experience reuse.
    \item We introduce \textbf{EvoNav-Bench}, a benchmark for LN in evolving environments.
    \item We benchmark UniGoal~\cite{Yin_2025_CVPR}, 3D-Mem~\cite{Yang_2025_CVPR} and MSGNav~\cite{huang2026msgnavunleashingpowermultimodal} on EvoNav-Bench and compare three simple heuristic strategies for handling environment evolution.
\end{itemize}


\section{Related Work}
\label{sec:related_work}

\textbf{Lifelong Navigation}\quad
In literature, LN can be formulated as a sequence of object navigation tasks.
MultiON~\cite{wani2020multion} represent the multiple objects with one-hot vectors for object categories.
GOAT-Bench~\cite{Khanna_2024_CVPR} extends this setting to multimodal goal specifications and provides agents with object categories, language descriptions, and images.
Meanwhile, another thread of work formulates LN as a continual learning problem: the learning phase consists of several stages, but in each episode there is only one goal object. 
C-Nav~\cite{NEURIPS2025_b78fef5a} instantiates this view as class incremental learning—agents gradually learn new target categories as learning proceeds, and LENL~\cite{wang2026lifelong} extends C-Nav to learning new scenes and instruction styles.
Despite these different formulations, these benchmarks assume static environments.
As a result, they cannot evaluate how environment evolution affects LN agents.

LN centers on experience consolidation, making navigation methods that build scene representations and long-term memories directly relevant.
Scene representation methods, such as 3D-Mem~\cite{Yang_2025_CVPR}, TANGO~\cite{Ziliotto_2025_CVPR}, MSGNav~\cite{huang2026msgnavunleashingpowermultimodal}, MTU3D~\cite{Zhu_2025_ICCV}, and MLFM~\cite{raychaudhuri2025mlfmmultilayeredfeaturemaps}, are applicable if the representation is made persistent across stages.
Persistent representations can take other forms as well, including topological graphs~\cite{Wiyatno_2022_RAL}, spatial-semantic graphs~\cite{niu2026ssmgnavenhancinglifelongobject}, self-refining graph memories~\cite{ji2025dynavlmzeroshotvisionlanguage}, or compressed visual contexts~\cite{ren2025astranavmemorycontextscompression}.
Similarly, methods that construct long-term trajectory memories~\cite{wang2026trajragretrievinggeometricsemanticexperience}, behavior memories~\cite{11459135}, or episodic-semantic memories~\cite{li2026himmhumaninspiredlongtermemory} are also applicable.
However, these methods overlook potential changes in the underlying environment, making them insufficient for real-world LN applications.
EvoNav-Bench provides an evaluation setting for studying LN agents under environment evolution.

\textbf{Dynamic and Lifelong SLAM}\quad 
Environmental changes have long been studied in SLAM. 
Dynamic SLAM handles moving objects during operation~\cite{1308004, Bescos_2018}, while lifelong SLAM address cross-session changes and persistent localization~\cite{shi2019openlorisscene,qian2023povslamprobabilisticobjectawarevariational}. These works primarily evaluate localization, mapping, or change estimation. EvoNav-Bench instead evaluates how unobserved between-stage changes affect cross-stage experience reuse in closed-loop ObjectNav.

\textbf{Related Benchmarks}\quad
A large body of non-lifelong navigation benchmarks studies how agents understand and ground diverse goals.
ObjectNav~\cite{batra2020objectnavrevisited} and HM3D-OVON~\cite{Yokoyama_2024_IROS} evaluate category-level and open-vocabulary object search, while LangNav~\cite{pan2024langnav}, SG3D~\cite{zhang2024taskorientedsequentialgrounding}, and LangMap~\cite{miao2026langmaphierarchicalbenchmarkopenvocabulary} extend navigation to language-specified, sequential, and hierarchical goals.
Vision-and-language navigation benchmarks analyze route-following perspective, spanning instruction following in R2R~\cite{Anderson_2018_CVPR}, remote object grounding in REVERIE~\cite{Qi_2020_CVPR}, and continuous-environment navigation in VLN-CE~\cite{Krantz_2020_ECCV}.
Closer to our setting, two recent benchmarks also introduce temporal changes into navigation and object search.
Portable ObjectNav~\cite{dorbala2024personalizedembodiednavigation} considers dynamic object placement in a given topological graph, whereas EvoNav-Bench involves exploration in continuous environment.
STARBench~\cite{chen2025searchingspacetimeunified} studies object retrieval in known environment with a given navigation policy, but EvoNav-Bench studies navigation in a unknown environment.


\section{EvoNav-Bench}
\label{sec:benchmark}

\subsection{Task Formulation}
\label{subsec:task-formulation}
Each task in EvoNav-Bench consists of 3--5 stages in a single unseen environment, and each stage is assigned its own step budget $T$.
The $k$\textsuperscript{th} stage prescribes a target object $x_k$ to be navigated to, described with its category, adjacent objects, appearance description, and an image.
A stage terminates either when the agent determines that the target has been reached or when its budget is exhausted; if another stage remains, the next target is then released regardless of the previous stage's success.
Starting from a random position, at each step $t$, an agent receives RGB-D observations $o_t$ and navigates either by taking discrete actions $a_t$ from \texttt{MOVE\_FORWARD}, \texttt{TURN\_LEFT}, and \texttt{TURN\_RIGHT} or heading towards a target position $p_t$.

In an EvoNav-Bench task, a subset of objects may be randomly relocated to feasible positions between consecutive stages after the current stage terminates.
No information of the object relocation is released to the agent.
This design simulates the case where human activities occur outside the robot's observation, leaving the agent to infer environment evolution only from later observations.
To isolate the challenge of lifelong navigation from online obstacle avoidance, we keep the environment stationary within each stage.
Under this formulation, agents are evaluated on navigation performance when environment evolution can make prior observations partially outdated.

\subsection{Task Generation}
\label{subsec:task-generation}
\textbf{Scenes and Tasks}\quad EvoNav-Bench is built on ProcTHOR-10k~\cite{NEURIPS2022_27c546ab}, a collection of procedurally generated indoor scenes with editable specifications.
Each ProcTHOR-10k scene contains one to ten rooms and assets drawn from 108 categories, providing diverse indoor layouts and object arrangements for object navigation.
More importantly, each scene is exposed as a structured configuration that specifies the floor plan, object instances, and object placements.
This editability allows EvoNav-Bench to generate controlled local scene evolution by modifying object placements within a scene.
We construct each EvoNav-Bench task from one ProcTHOR-10k scene and filter out single-room scenes so that scene evolution can occur outside the agent's current room.
For each retained scene, we generate a 3--5-stage object-navigation task, following the principles and details described below.

\textbf{Scene Evolution Principles}\quad Recall that EvoNav-Bench targets the LN scenario in which accumulated experience is \textit{useful but potentially unreliable}.
This goal motivates four design dimensions for scene evolution.
For \textbf{locality}, each evolution is limited to one selected room, leaving the rest of the scene unchanged so that majority of accumulated experience can remain useful.
For \textbf{experience relevance}, affected objects are chosen from previously explored areas, allowing outdated object locations or local layout cues to conflict with later observations.
For \textbf{observability}, evolution occurs outside the agent's current observation, so changes must be inferred later rather than directly observed.
Finally, \textbf{post-evolution validity} requires the evolved scene to remain physically plausible and navigable. Specifically, AI2-THOR's shortest-path planner are used to verify the reachability of goals after evolution.

\begin{figure}
    \centering
    \includegraphics[width=1.0\textwidth]{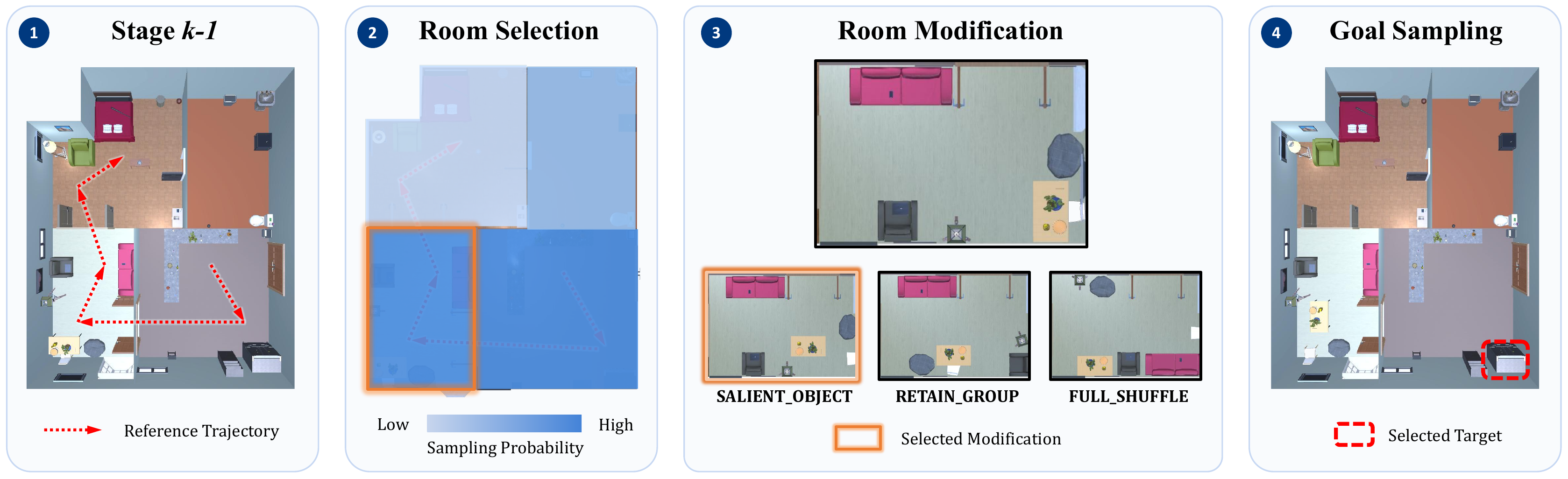}
    \caption{\textbf{Scene evolution and goal sampling from stage $k-1$ to stage $k$ in EvoNav-Bench.}
    \textbf{(1) Reference trajectory:} during task generation, a reference trajectory is used to determine visited rooms and the transition room before stage $k$.
    \textbf{(2) Room selection:} a room is sampled using room assets, visited rooms, the transition room, and recent evolution history.
    \textbf{(3) Room modification:} one of three modification approaches is applied to the selected room.
    \texttt{SALIENT\_OBJECT} relocates a large object, such as the yellow desk in the figure, together with objects placed on it;
    \texttt{RETAIN\_GROUP} relocates all objects in the room while retaining asset groups, such as the desk and the white chair;
    and \texttt{FULL\_SHUFFLE} relocates all objects independently.
    \textbf{(4) Goal sampling:} after room modification, the target object for stage $k$ is sampled, and it may or may not reside in the changed room.}
    \label{fig:evolve-pipeline}
    \vspace{-1.5em}
\end{figure}

\textbf{Task Generation Details}\quad EvoNav-Bench runs task generation and agent evaluation stage by stage.
The generation process produces a stage-wise evolution schedule and the corresponding scene configurations for each task, which are used for evaluation.
During evaluation, after each stage terminates by success or budget exhaustion, EvoNav-Bench saves the agent's current pose, loads the scene configuration for the next stage, and re-spawns the agent at the saved pose.

Constructing this schedule requires selecting which room, if any, should evolve at each stage transition.
Since the decision should not rely on any particular agent rollout, we use a \textit{reference trajectory} during task generation, as shown in the first panel of Fig.~\ref{fig:evolve-pipeline}.
The reference trajectory is generated stage by stage.
At stage $k=1$, we use the initial position of the corresponding ProcTHOR house and sample the first goal, then compute the first trajectory segment from the initial position to the goal using the \texttt{GetShortestPathToPoint} function in AI2-THOR.
For each later stage, after scene evolution and goal sampling, we append a shortest-path segment from the previous reference endpoint to the new goal.
The accumulated trajectory is used only during task generation to determine visited rooms and the transition room, which guide subsequent scene evolution and goal sampling.

We next describe how the room to be modified is sampled, which is illustrated in the second panel of Fig.~\ref{fig:evolve-pipeline}.
The dataset-generation parameters are fixed to balance three properties: environment evolution should occur frequently enough to affect multi-stage navigation, changes should remain local, and goals should be related to evolved regions without always being relocated objects.
For each transition from stage $k-1$ to stage $k$, scene evolution is triggered with probability $p^\text{evolve}=0.5$.
When evolution is triggered, we sample a room from a categorical distribution:
\begin{equation}
    p^\text{evolved\ room}_j\propto q^\text{asset}_jq^\text{visited}_jq^\text{current}_jq^\text{evolved}_j.
\end{equation}
The asset term $q^\text{asset}_j$ is the number of assets in the $j$\textsuperscript{th} room, encouraging evolution in rooms with richer object layouts.
Let $\mathbb{I}(\cdot)$ denote the indicator function.
The other terms use the reference trajectory and recent evolution history to favor previously visited rooms, exclude the transition room, and avoid repeatedly modifying the same room:
\begin{equation}
    \begin{split}
        q^\text{visited}_j&=f^\text{visited}\mathbb{I}(\text{room\ }j\text{\ has\ been\ visited})+\left(1-\mathbb{I}(\text{room\ }j\text{\ has\ been\ visited})\right) \\
        q^\text{current}_j&=1-\mathbb{I}(\text{room\ }j\text{\ is\ the\ transition room}) \\
        q^\text{evolved}_j&=f^\text{evolved}\mathbb{I}(\text{room\ }j\text{\ was\ sampled\ at\ }k-2)+\left(1-\mathbb{I}(\text{room\ }j\text{\ was\ sampled\ at\ }k-2)\right).
    \end{split}
\end{equation}
We set $f^\text{visited}=3$ and $f^\text{evolved}=0.5$, with all constants fixed before agent evaluation to define the benchmark distribution.
Together, this room-sampling mechanism implements \textbf{locality}, \textbf{experience relevance}, and \textbf{observability}, while reducing repeated evolution of the same room across consecutive stages.

After room selection, we apply one of three room modification approaches, as shown in the third panel of Fig.~\ref{fig:evolve-pipeline}.
\texttt{SALIENT\_OBJECT} relocates one salient object together with smaller objects placed on it, whereas \texttt{RETAIN\_GROUP} and \texttt{FULL\_SHUFFLE} relocate all objects in the room while either preserving semantic groups or moving objects independently.
All relocation positions are selected with the ProcTHOR placement algorithm to ensure navigability.

\textbf{Goal Sampling}\quad We sample navigation goals to make them both unambiguous and likely to reflect the influence of scene evolution, a step illustrated in the goal-sampling panel of Fig.~\ref{fig:evolve-pipeline}.
To reduce goal ambiguity, we define a \textit{target score} for each object based on its uniqueness.
The score is 2 if the object is unique in the entire scene, 1.5 if it is unique within its room but not in the entire scene, and 1 otherwise.
We use these scores to sample goals hierarchically, first sampling a goal room from a categorical distribution.
The weight of each room is the sum of target scores of objects in that room, multiplied by 9 if the room is modified for the current stage and by 0.2 if it is the same goal room as in the previous stage.
This biases goals toward evolved rooms while discouraging consecutive stages from selecting the same goal room, but it does not force the target to be a relocated object.
Within the sampled goal room, candidate objects are sampled uniformly by default.
If the current stage uses \texttt{SALIENT\_OBJECT}, the moved object is assigned weight 9, further increasing the chance that the goal is affected by scene evolution.

\begin{figure}
    \centering
    \includegraphics[width=1.0\textwidth]{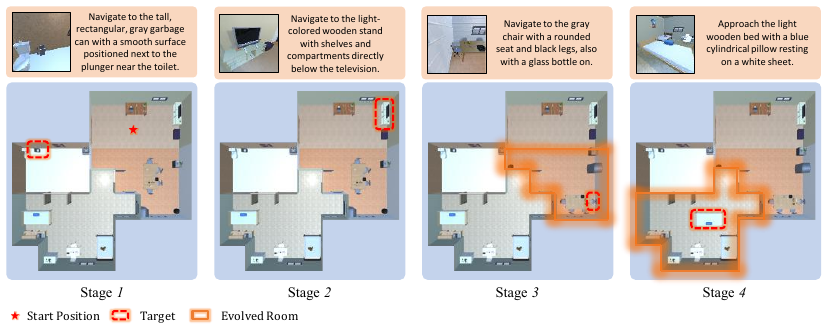}
    \caption{\textbf{An Example task in EvoNav-Bench.}
    This task occurs in a house with four rooms.
    The goal of each stage is described with an image and a short description consisting of category, adjacent objects, and appearance description.
    At stage 3, the bottom left room is modified with \texttt{RETAIN\_GROUP} method. At stage 4, the bottom left room is changed with \texttt{SALIENT\_OBJECT} method.}
    \label{fig:example}
    \vspace{-1.5em}
\end{figure}

\begin{figure}[t]
    \centering
    \begin{subfigure}[t]{0.3\textwidth}
        \centering
        \includegraphics[width=\linewidth]{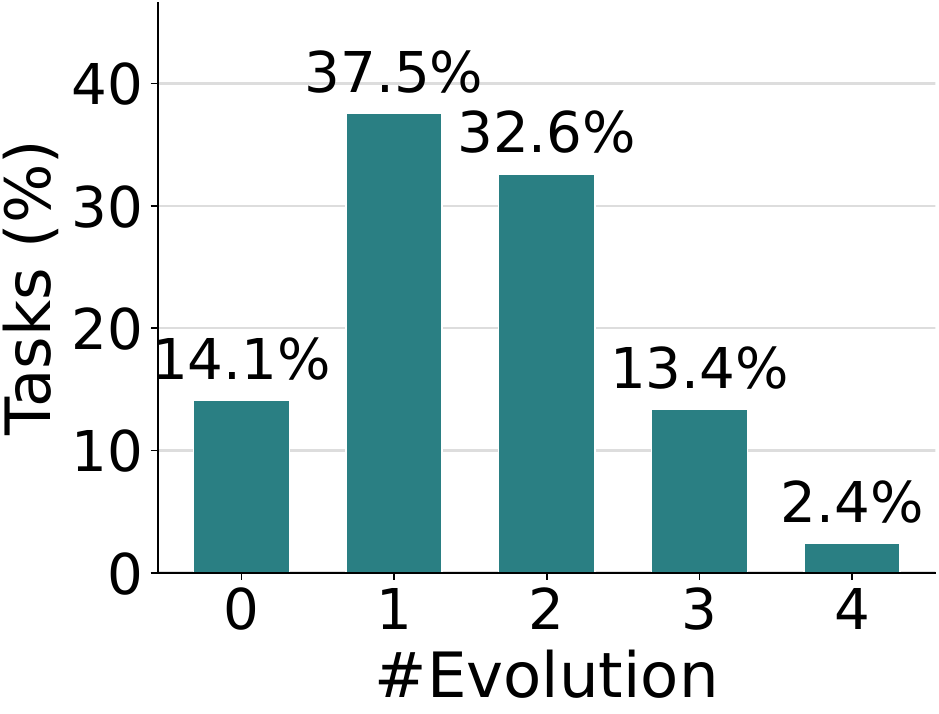}
        \caption{Environment evolution count.}
        \label{fig:stats-stage-changes}
    \end{subfigure}
    \hfill
    \begin{subfigure}[t]{0.3\textwidth}
        \centering
        \includegraphics[width=\linewidth]{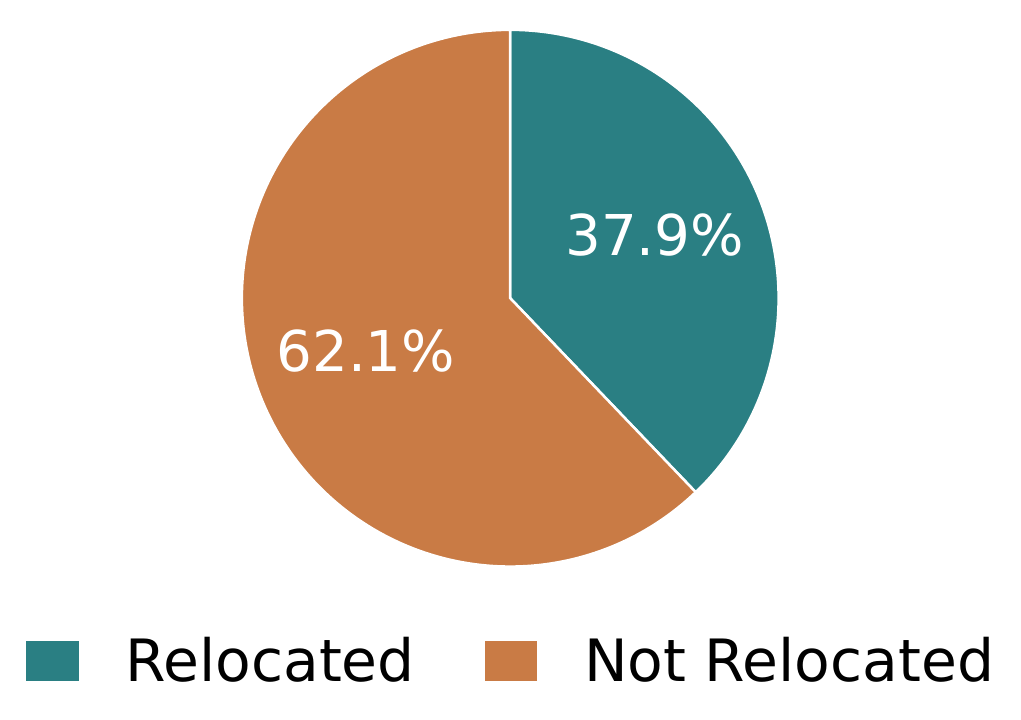}
        \caption{Stage-goal objects.}
        \label{fig:stats-target-position-history}
    \end{subfigure}
    \hfill
    \begin{subfigure}[t]{0.38\textwidth}
        \centering
        \includegraphics[width=\linewidth]{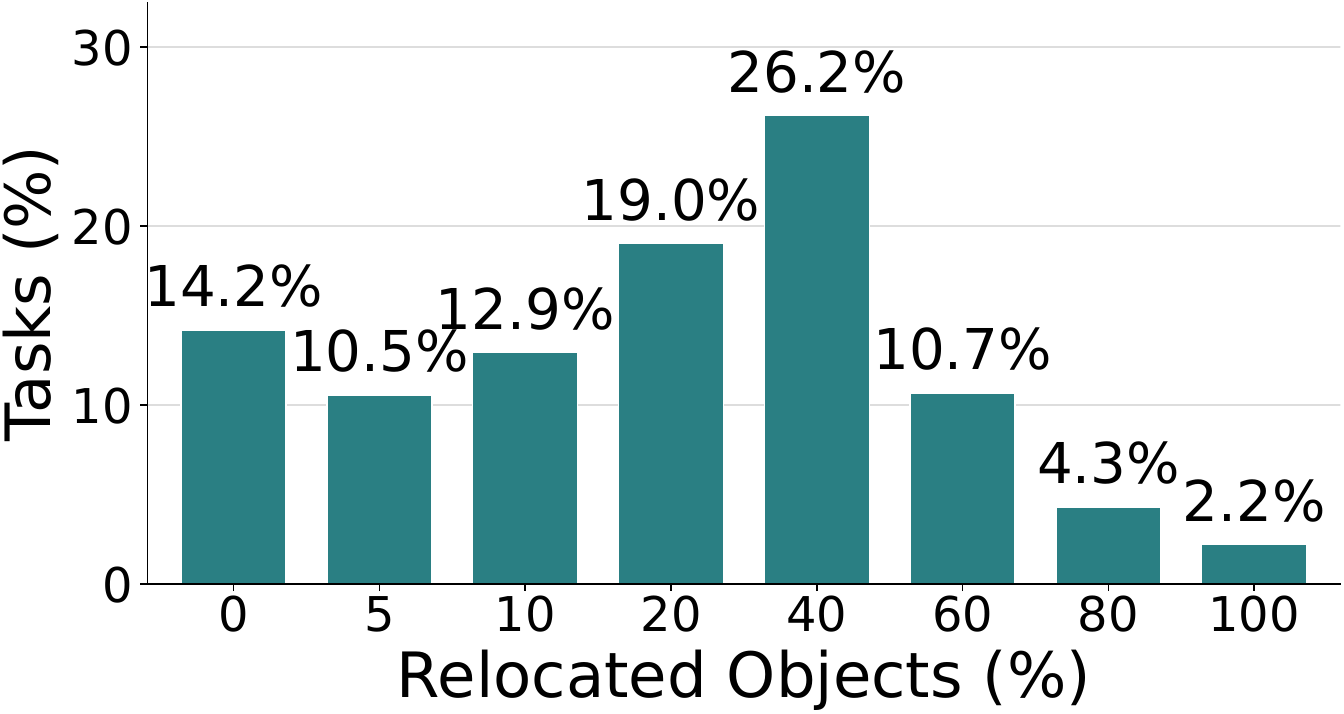}
        \caption{Relocated-object ratio.}
        \label{fig:stats-moved-object-ratio}
    \end{subfigure}
    \caption{\textbf{Environment-evolution statistics in EvoNav-Bench.}
    Scene evolution occurs in 85.9\% of episodes, with 48.4\% containing two or more evolved stage transitions.
    Among stage-goal objects, 37.9\% have been relocated before being assigned as targets.
    Across episodes, the relocated-object ratio spans from no relocation to over 80\% of house objects, with median value within 10\%-20\%, showing that EvoNav-Bench covers both mild and substantial scene changes.}
    \label{fig:benchmark-stats}
    \vspace{-1.0em}
\end{figure}

\textbf{Statistics and Examples}\quad Fig.~\ref{fig:example} shows an example task in EvoNav-Bench.
EvoNav-Bench inherits its training, validation, and test splits from the ProcTHOR-10k dataset.
After filtering out single-room scenes, we generate 7,924 training tasks with 31,797 stages, 751 validation tasks with 3,003 stages, and 776 test tasks with 3,098 stages.
As shown in Fig.~\ref{fig:stats-stage-changes}, scene evolution occurs at least once in 85.9\% of tasks and two or more times in 48.4\% of tasks, indicating that agents frequently face evolving environments. Nevertheless, 49.4\% of the stages are static.
At the same time, Fig.~\ref{fig:stats-target-position-history} shows that 62.1\% of stage goals have not been relocated in previous stages, indicating that prior experience can still provide useful navigation cues.
Finally, Fig.~\ref{fig:stats-moved-object-ratio} shows that the relocated-object ratio ranges from 0\% to over 80\%, with a median below 20\%, indicating that EvoNav-Bench mostly produces localized scene evolution.
Together, these statistics indicate that the generation parameters do not make EvoNav-Bench a relocated-object search benchmark: most tasks contain environment evolution, but most stage goals have not themselves been relocated, and scene changes remain localized in typical episodes.


\section{Experiments}
\label{sec:experiments}
We use EvoNav-Bench to evaluate whether agents that reuse accumulated scene representations remain reliable when scenes evolve between stages.
We compare three representative agents: UniGoal, 3D-Mem, and MSGNav. 
We also design three simple heuristics: Frontier-Update, Fail-then-Update and Stage-Reset.
These experiments characterize how prior experience affect performance under scene evolution.

\subsection{Experimental Setup}
\label{sec:experimental-setup}

\textbf{Data and Metrics}\quad
For computational feasibility, we evaluate agents on a 10\% subset of the EvoNav-Bench test split, containing 90 unique scenes and 365 stages.
This subset remains on the same order of magnitude as the GOAT-Bench experiments reported in~\cite{Yang_2025_CVPR}, which use 36 scenes and 278 navigation subtasks.
We report success rate (SR) and success-weighted path length (SPL) at the stage level.
SR measures whether the agent reaches the target object within the stage budget, while SPL weights successful stages by path efficiency, using the shortest path to the stage target as reference.
We aggregate both metrics over stages.
To isolate the influence of perception quality, we evaluate all methods under two settings: \textbf{GT} setting, where object masks and semantic labels are provided by ground truth, and \textbf{Prediction} setting, where they are produced by vision models.

\textbf{Compared Methods}\quad
We compare three recent agents with persistent scene representations.
\textbf{UniGoal~\cite{Yin_2025_CVPR}} builds an online scene graph, converts category, image, and language goals into object-relation goal graphs, and uses graph matching with LLM reasoning for exploration and target verification.
\textbf{3D-Mem~\cite{Yang_2025_CVPR}} represents explored regions as visual snapshots and unexplored regions as frontier snapshots, then uses a VLM to select relevant snapshots for navigation and exploration.
\textbf{MSGNav~\cite{huang2026msgnavunleashingpowermultimodal}} builds a multi-modal 3D scene graph, introduces several mechanisms to handle perception error, and a VLM is used for navigation, exploration and target verification.

We also provides three simple heuristics for handling environment evolution and diagnosis.
Specifically, the \textbf{Frontier-Update (FU)} method maintains accumulated object instances, an occupancy map across stages, and frontiers extracted from occupancy map.
When current observations conflict with the current occupancy map, e.g., when a previously occupied cell becomes free, the occupancy map is rebuilt from subsequent observations and all exploration frontiers are reset.
This keeps exploration frontiers consistent with the evolved environment.
\textbf{Fail-then-Update (FTU)} is built upon FU, checking whether the target is present upon reaching its stored location and revising the entry only if the target is absent. Once the stale instance is revised, the agent continues to find the target.
\textbf{Stage-Reset} clears both FU's object entries and occupancy map at the beginning of each stage, which means the agent searches from scratch. Consequently, it is not affected by stale object entries, but it also cannot leverage any prior experience.

\textbf{Implementation Details}\quad For UniGoal, 3D-Mem and MSGNav, we use their original implementations, but the occupancy map is provided by the simulator.
All heuristics share the same pipeline for constructing ConceptGraph-like object instances, using the occupancy map from simulator, generating frontiers, and executing navigation.
Under Prediction setting, this shared pipeline uses the same vision models as in the case of 3D-Mem, consisting of YOLO-World~\cite{Cheng_2024_CVPR} to detect object boxes, SAM~\cite{Kirillov_2023_ICCV} to predict object masks, and OpenCLIP~\cite{Cherti_2023_CVPR} to associate new detections with stored object entries.
The agent's final pose from the previous stage may lie in the evolved room. We preserve it if it remains navigable; otherwise, we move the agent to the nearest navigable position to ensure continuity.

\subsection{Results}
\textbf{Evaluated baselines are brittle under scene evolution, heuristics are more effective.} As shown in Tab.~\ref{tab:main-results}, all heuristics outperform the evaluated baselines under GT setting, suggesting that baselines built without considering environment evolution are vulnerable to scene changes. Despite maintaining accumulated object entries, FU performs better considering that it adapts exploration frontiers to the evolved environment, which allows re-exploration in the evolved scene. Stage-Reset performs comparably to FU without maintaining cross-stage representations. However, since it explores from scratch at each stage, it achieves lower SPL than FU. FTU performs best among the methods, suggesting the reactive correction strategy that revising stale object entries is effective in handling scene evolution.

\textbf{Perception error can outweigh the influence of environment evolution.} Under Prediction setting, MSGNav which considers perception error performs best, suggesting that perception error can outweigh the influence of environment evolution. 3D-Mem can outperform FU with its robust snapshot based approach, and Stage-Reset can also outperform FU since it does not accumulate cross-stage object entries that can be wrong or stale. The performance of agents can be influenced by both erroneous object entries caused by perception error and stale object entries caused by scene evolution. 

\textbf{Impact of target relocation.} We split stages that may involve scene evolution by whether the target object is relocated as shown in Tab.~\ref{tab:relocation-results}. When the target is relocated, all methods suffer a performance drop. In the GT setting, Stage-Reset improves performance on relocated targets over FU but drops on non-relocated targets, exposing the reset-reuse trade-off. FTU improves performance on relocated targets while remaining close to FU on non-relocated targets, suggesting that reactive correction mainly recovers failures caused by target relocation. FTU still performs best on relocated targets in the Prediction setting, indicating that memory correction strategy is beneficial for handling scene evolution.

\begin{table}[!t]
\centering
\caption{\textbf{Main Results of evaluated baselines and heuristic strategies.}
Heuristics performs better than evaluated baselines under GT setting, while MSGNav which considers perception error performs best under Prediction setting.
}
\label{tab:main-results}
\small
\begin{tabular}{@{}llcccccc@{}}
\toprule
\multirow{2}{*}{\textbf{Setting}} & \multirow{2}{*}{\textbf{Metric}}
& \multicolumn{3}{c}{\textbf{Evaluated Baselines}}
& \multicolumn{3}{c}{\textbf{Heuristic Strategies}} \\
\cmidrule(lr){3-5}\cmidrule(lr){6-8}
& & UniGoal & 3D-Mem & MSGNav & FU & Stage-Reset & FTU \\
\midrule
\multirow{2}{*}{\textbf{GT}}
& SR $\uparrow$ & 20.0 & 47.7 & 59.2 & 61.1 & 61.9 & \textbf{67.4} \\
& SPL $\uparrow$ & 14.9 & 42.5 & 41.3 & 48.8 & 45.1 & \textbf{53.3} \\
\midrule
\multirow{2}{*}{\textbf{Prediction}}
& SR $\uparrow$ & 12.9 & 44.1 & \textbf{54.8} & 42.2 & 46.8 & 49.0 \\
& SPL $\uparrow$ & 7.9 & 37.7 & \textbf{38.2} & 32.5 & 29.6 & 35.8 \\
\bottomrule
\end{tabular}
\vspace{-1.0em}
\end{table}

\begin{table}[!t]
\centering
\caption{\textbf{Impact of target relocation on navigation performance.}
We report the success rate (SR) of evaluated methods, split by whether the target object is relocated. All methods suffer a performance drop on relocated targets, and FTU performs best.
}
\label{tab:relocation-results}
\small
\begin{tabular}{llcccccc}
\toprule
\textbf{Perception} & \textbf{Target Status} & \textbf{UniGoal} & \textbf{3D-Mem} & \textbf{MSGNav} & \textbf{FU} & \textbf{Stage-Reset} & \textbf{FTU} \\
\midrule
\multirow{2}{*}{\textbf{GT}}
& Relocated     & 12.8 & 40.4 & 51.4 & 54.1 & 60.6 & \textbf{68.8} \\
& Not Relocated & 18.7 & 56.0 & 64.5 & 65.1 & 61.4 & \textbf{66.3} \\
\midrule
\multirow{2}{*}{\textbf{Prediction}}
& Relocated     & 9.2 & 40.4 & 50.7 & 29.4 & 44.0 & \textbf{52.3} \\
& Not Relocated & 10.8 & 48.1 & \textbf{57.8} & 50.0 & 49.4 & 47.9 \\
\bottomrule
\end{tabular}
\vspace{-1.0em}
\end{table}

\subsection{Controlled Analysis for Environment Evolution}
\label{sec:controlled-analysis}

We conduct a controlled analysis to isolate the influence of environment evolution from accumulated experience in previous stage and perception quality.

\textbf{Setup}\quad
We evaluate 90 ProcTHOR-10k test scenes under GT perception.
For each scene, agents are first exposed with Stage-1 and Stage-2 goals. Then, agents are forced to navigate to the Stage-2 goal, then navigate to the Stage-1 goal, following the AI2-THOR's shortest-path planner. Concurrently, agents receive observations along the trajectory and update their scene representations.
We then evaluate the same Stage-2 goal under two variants: \textbf{Fixed}, where the scene remains unchanged, and \textbf{Not Fixed}, where the scene evolves between stages following Sec.~\ref{subsec:task-generation}.
We report second-stage results for 3D-Mem, MSGNav, FU, Stage-Reset, and FTU, and omit UniGoal due to its poor performance.

\begin{figure}[t]
    \centering
    \includegraphics[width=.8\textwidth]{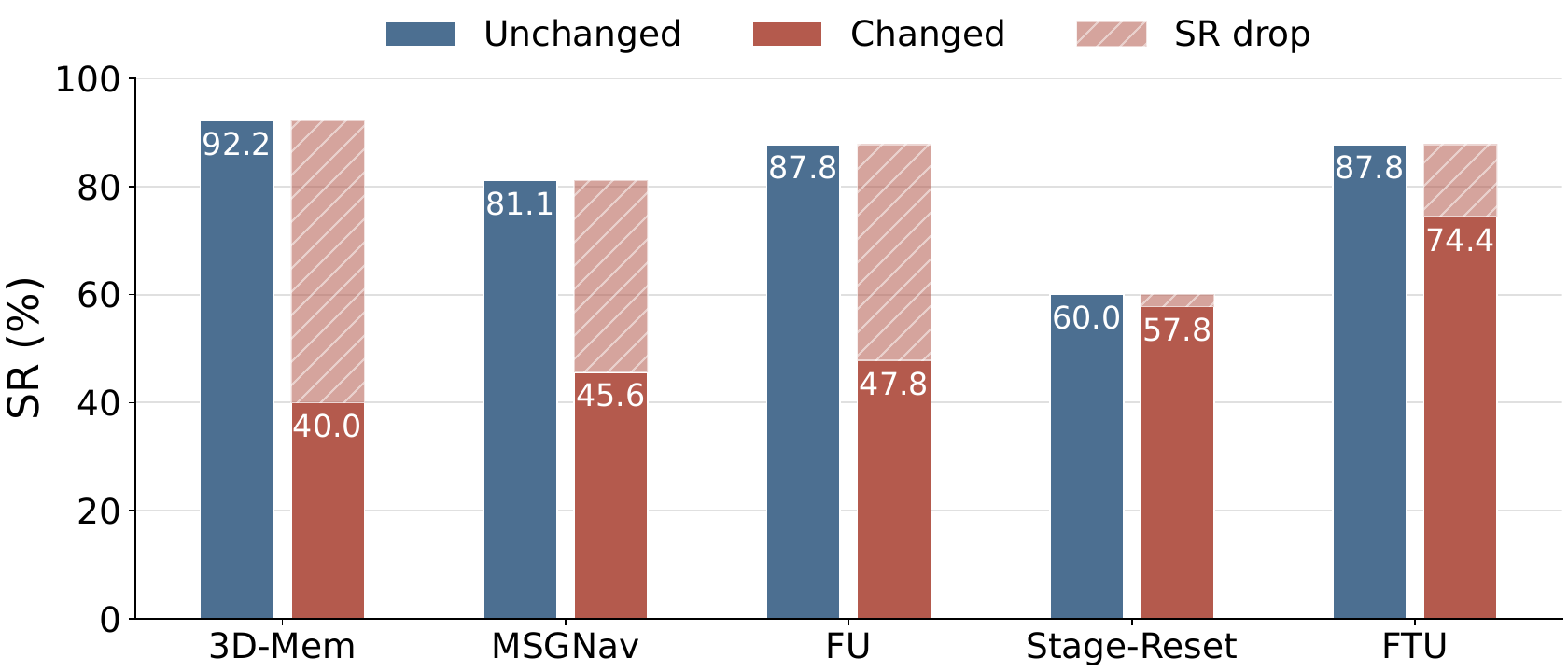}
    \caption{\textbf{Retained object entries help in fixed scenes but can mislead navigation after scene evolution.}
    Performance drops after scene evolution.
    FU outperforms Stage-Reset when the environment evolution is disabled, but drops below Stage-Reset after scene evolution.
    MSGNav and FTU drops relatively less despite maintaining object entries.
    .}
    \label{fig:gt-exp2}
    \vspace{-1.0em}
\end{figure}

\textbf{Retained object entries can become harmful after scene evolution.}
Experiment setting allow the agent to construct entry of Stage-2 goal in advance. Fig.~\ref{fig:gt-exp2} shows that methods maintaining object entries (3D-Mem, MSGNav, FU, FTU) outperform Stage-Reset when the environment is Fixed, which demonstrates the benefit of reusing accumulated scene representations. However, when the environment evolves, these methods drop significantly, while Stage-Reset maintains it performance with 2.2\% SR drop, suggesting that retained object entries can mislead navigation after scene evolution. 

\textbf{Target verification and reactive correction strategy can mitigate the negative impact of stale object entries.}
MSGNav and FTU drop relatively less than 3D-Mem and FU. MSGNav equips a target verification module to check whether the target is reached, which can somehow prevent the mismatch between the stored representation and the current scene. FTU revises object entries after a verification on the staleness of the chosen target. FTU performs better since it can revise the stale object entries but MSGNav still maintains these entries. This suggests that target verification and reactive correction strategy can mitigate the negative impact of stale object entries. Handling staleness is the key problem in lifelong navigation with evolving environments. 


\section{Conclusion}
\label{sec:conclusion}


We introduce EvoNav-Bench, a benchmark for evaluating lifelong navigation in environments that evolve between navigation stages.
Through controlled object relocation without notifying the agent, EvoNav-Bench creates a setting in which accumulated scene representations remain useful but can become outdated.
Evaluations of three recent navigation agents and three diagnostic heuristic strategies show that methods relying on persistent scene representations can be brittle under environment evolution.
The main results expose a trade-off between reusing prior information and re-exploring changed scenes, while the predicted-perception setting shows that perception errors can outweigh the influence of environment evolution.
The controlled Fixed/Not Fixed analysis further isolates the effect of stale object entries: retaining entries improves navigation when the environment remains unchanged, but can substantially degrade performance after evolution. Reactive entry correction can mitigate this degradation.
Together, these findings suggest that lifelong navigation in evolving environments requires agents to balance experience reuse with the revision of stale information and perception errors.

\textbf{Limitations} \quad EvoNav-Bench currently models environment evolution through simulated object relocation, which may not fully reflect real-world rearrangements and does not cover object insertion and removal, articulated-state changes, or object-relocation strategies from human activity data.
While ProcTHOR enables controlled scene generation, its simulated environments are less realistic than real-world settings. Future work can migrate EvoNav-Bench to higher-fidelity simulators such as Isaac Sim or adopt Real2Sim pipelines. 
The evaluated update strategies are also reactive and remain sensitive to prediction errors.
Future work can extend EvoNav-Bench with richer evolution mechanisms and develop LN agents with proactive strategies that perform reliably in evolving environments.


\clearpage


\bibliography{references}  

@InProceedings{Khanna_2024_CVPR,
    author    = {Khanna, Mukul and Ramrakhya, Ram and Chhablani, Gunjan and Yenamandra, Sriram and Gervet, Theophile and Chang, Matthew and Kira, Zsolt and Chaplot, Devendra Singh and Batra, Dhruv and Mottaghi, Roozbeh},
    title     = {GOAT-Bench: A Benchmark for Multi-Modal Lifelong Navigation},
    booktitle = {Proceedings of the IEEE/CVF Conference on Computer Vision and Pattern Recognition (CVPR)},
    year      = {2024},
    pages     = {16373-16383}
}

@InProceedings{Yang_2025_CVPR,
    author    = {Yang, Yuncong and Yang, Han and Zhou, Jiachen and Chen, Peihao and Zhang, Hongxin and Du, Yilun and Gan, Chuang},
    title     = {3D-Mem: 3D Scene Memory for Embodied Exploration and Reasoning},
    booktitle = {Proceedings of the IEEE/CVF Conference on Computer Vision and Pattern Recognition (CVPR)},
    year      = {2025},
    pages     = {17294-17303}
}

@InProceedings{Yin_2025_CVPR,
    author    = {Yin, Hang and Xu, Xiuwei and Zhao, Linqing and Wang, Ziwei and Zhou, Jie and Lu, Jiwen},
    title     = {UniGoal: Towards Universal Zero-shot Goal-oriented Navigation},
    booktitle = {Proceedings of the IEEE/CVF Conference on Computer Vision and Pattern Recognition (CVPR)},
    month     = {June},
    year      = {2025},
    pages     = {19057-19066},
    doi       = {10.1109/CVPR52734.2025.01775}
}

@InProceedings{Ziliotto_2025_CVPR,
    author    = {Ziliotto, Filippo and Campari, Tommaso and Serafini, Luciano and Ballan, Lamberto},
    title     = {TANGO: Training-free Embodied AI Agents for Open-world Tasks},
    booktitle = {Proceedings of the IEEE/CVF Conference on Computer Vision and Pattern Recognition (CVPR)},
    year      = {2025},
    pages     = {24603-24613}
}

@misc{huang2026msgnavunleashingpowermultimodal,
      title={MSGNav: Unleashing the Power of Multi-modal 3D Scene Graph for Zero-Shot Embodied Navigation},
      author={Xun Huang and Shijia Zhao and Yunxiang Wang and Xin Lu and Wanfa Zhang and Rongsheng Qu and Weixin Li and Yunhong Wang and Chenglu Wen},
      year={2026},
      eprint={2511.10376},
      archivePrefix={arXiv},
      primaryClass={cs.CV},
}

@misc{wang2026trajragretrievinggeometricsemanticexperience,
      title={TrajRAG: Retrieving Geometric-Semantic Experience for Zero-Shot Object Navigation},
      author={Yiyao Wang and Sixian Zhang and Keming Zhang and Xinhang Song and Songjie Du and Shuqiang Jiang},
      year={2026},
      eprint={2605.01700},
      archivePrefix={arXiv},
      primaryClass={cs.CV},
}

@ARTICLE{11459135,
  author={Xu, Yunzhe and Pan, Yiyuan and Liu, Zhe},
  journal={IEEE Transactions on Pattern Analysis and Machine Intelligence},
  title={Dream to Recall: Imagination-Guided Experience Retrieval for Memory-Persistent Vision-and-Language Navigation},
  year={2026},
  doi={10.1109/TPAMI.2026.3679426}
}

@inproceedings{
wang2026lifelong,
title={Lifelong Embodied Navigation Learning},
author={Xudong Wang and Jiahua Dong and Baichen Liu and Qi Lyu and Lianqing Liu and Zhi Han},
booktitle={The Fourteenth International Conference on Learning Representations},
year={2026}
}

@inproceedings{NEURIPS2025_b78fef5a,
 author = {Yu, MingMing and Zhu, Fei and Liu, Wenzhuo and Yang, Yirong and Wang, Qunbo and wu, wenjun and Liu, Jing},
 booktitle = {Advances in Neural Information Processing Systems},
 editor = {D. Belgrave and C. Zhang and H. Lin and R. Pascanu and P. Koniusz and M. Ghassemi and N. Chen},
 pages = {126181--126207},
 publisher = {Curran Associates, Inc.},
 title = {C-NAV: Towards Self-Evolving Continual Object Navigation in Open World},
 volume = {38},
 year = {2025}
}

@inproceedings{Churchill_2012_ICRA,
  author    = {Churchill, Winston and Newman, Paul},
  title     = {Practice Makes Perfect? Managing and Leveraging Visual Experiences for Lifelong Navigation},
  booktitle = {2012 IEEE International Conference on Robotics and Automation},
  year      = {2012},
  pages     = {4525--4532},
  doi       = {10.1109/ICRA.2012.6224596}
}

@inproceedings{wani2020multion,
  author    = {Wani, Saim and Patel, Shivansh and Jain, Unnat and Chang, Angel and Savva, Manolis},
  title     = {MultiON: Benchmarking Semantic Map Memory using Multi-Object Navigation},
  booktitle = {Advances in Neural Information Processing Systems},
  year      = {2020},
  volume    = {33},
  pages     = {9700--9712}
}

@inproceedings{Chang_RSS_24,
  author    = {Chang, Matthew and Gervet, Theophile and Khanna, Mukul and Yenamandra, Sriram and Shah, Dhruv and Min, So Yeon and Shah, Kavit and Paxton, Chris and Gupta, Saurabh and Batra, Dhruv and Mottaghi, Roozbeh and Malik, Jitendra and Chaplot, Devendra Singh},
  title     = {{GOAT: GO to Any Thing}},
  booktitle = {Proceedings of Robotics: Science and Systems},
  year      = {2024},
  address   = {Delft, Netherlands},
  month     = {July},
  doi       = {10.15607/RSS.2024.XX.073}
}

@inproceedings{NEURIPS2022_27c546ab,
  author    = {Deitke, Matt and VanderBilt, Eli and Herrasti, Alvaro and Weihs, Luca and Salvador, Jordi and Ehsani, Kiana and Han, Winson and Kolve, Eric and Farhadi, Ali and Kembhavi, Aniruddha and Mottaghi, Roozbeh},
  title     = {{ProcTHOR}: Large-Scale Embodied AI Using Procedural Generation},
  booktitle = {Advances in Neural Information Processing Systems},
  year      = {2022},
  volume    = {35},
  pages     = {5982--5994},
  doi       = {10.52202/068431-0433}
}

@article{Wiyatno_2022_RAL,
  author  = {Wiyatno, Rey Reza and Xu, Anqi and Paull, Liam},
  title   = {Lifelong Topological Visual Navigation},
  journal = {IEEE Robotics and Automation Letters},
  year    = {2022},
  volume  = {7},
  number  = {4},
  pages   = {9271--9278},
  doi     = {10.1109/LRA.2022.3189164}
}

@misc{niu2026ssmgnavenhancinglifelongobject,
  title         = {{SSMG-Nav}: Enhancing Lifelong Object Navigation with Semantic Skeleton Memory Graph},
  author        = {Haochen Niu and Lantao Zhang and Xingwu Ji and Rendong Ying and Peilin Liu and Fei Wen},
  year          = {2026},
  eprint        = {2603.01813},
  archivePrefix = {arXiv},
  primaryClass  = {cs.RO}
}

@misc{ren2025astranavmemorycontextscompression,
  title         = {{AstraNav-Memory}: Contexts Compression for Long Memory},
  author        = {Botao Ren and Junjun Hu and Xinda Xue and Minghua Luo and Jintao Chen and Haochen Bai and Liangliang You and Mu Xu},
  year          = {2025},
  eprint        = {2512.21627},
  archivePrefix = {arXiv},
  primaryClass  = {cs.RO}
}

@InProceedings{Zhu_2025_ICCV,
  author    = {Zhu, Ziyu and Wang, Xilin and Li, Yixuan and Zhang, Zhuofan and Ma, Xiaojian and Chen, Yixin and Jia, Baoxiong and Liang, Wei and Yu, Qian and Deng, Zhidong and Huang, Siyuan and Li, Qing},
  title     = {Move to Understand a 3D Scene: Bridging Visual Grounding and Exploration for Efficient and Versatile Embodied Navigation},
  booktitle = {Proceedings of the IEEE/CVF International Conference on Computer Vision (ICCV)},
  month     = {October},
  year      = {2025},
  pages     = {8120--8132}
}

@misc{raychaudhuri2025mlfmmultilayeredfeaturemaps,
  title         = {{MLFM}: Multi-Layered Feature Maps for Richer Language Understanding in Zero-Shot Semantic Navigation},
  author        = {Sonia Raychaudhuri and Enrico Cancelli and Tommaso Campari and Lamberto Ballan and Manolis Savva and Angel X. Chang},
  year          = {2025},
  eprint        = {2507.07299},
  archivePrefix = {arXiv},
  primaryClass  = {cs.RO}
}

@misc{ji2025dynavlmzeroshotvisionlanguage,
  title         = {{DyNaVLM}: Zero-Shot Vision-Language Navigation System with Dynamic Viewpoints and Self-Refining Graph Memory},
  author        = {Zihe Ji and Huangxuan Lin and Yue Gao},
  year          = {2025},
  eprint        = {2506.15096},
  archivePrefix = {arXiv},
  primaryClass  = {cs.RO}
}

@misc{li2026himmhumaninspiredlongtermemory,
  title         = {{HIMM}: Human-Inspired Long-Term Memory Modeling for Embodied Exploration and Question Answering},
  author        = {Ji Li and Bo Wang and Jing Xia and Mingyi Li and Shiyan Hu},
  year          = {2026},
  eprint        = {2602.15513},
  archivePrefix = {arXiv},
  primaryClass  = {cs.RO}
}

@inproceedings{Yokoyama_2024_IROS,
  author    = {Yokoyama, Naoki and Ramrakhya, Ram and Das, Abhishek and Batra, Dhruv and Ha, Sehoon},
  title     = {{HM3D-OVON}: A Dataset and Benchmark for Open-Vocabulary Object Goal Navigation},
  booktitle = {2024 IEEE/RSJ International Conference on Intelligent Robots and Systems (IROS)},
  year      = {2024},
  pages     = {5543--5550},
  doi       = {10.1109/IROS58592.2024.10802709}
}

@misc{batra2020objectnavrevisited,
  title         = {{ObjectNav} Revisited: On Evaluation of Embodied Agents Navigating to Objects},
  author        = {Dhruv Batra and Aaron Gokaslan and Aniruddha Kembhavi and Oleksandr Maksymets and Roozbeh Mottaghi and Manolis Savva and Alexander Toshev and Erik Wijmans},
  year          = {2020},
  eprint        = {2006.13171},
  archivePrefix = {arXiv},
  primaryClass  = {cs.CV}
}

@misc{zhang2024taskorientedsequentialgrounding,
  title         = {Task-oriented Sequential Grounding and Navigation in 3D Scenes},
  author        = {Zhuofan Zhang and Ziyu Zhu and Junhao Li and Pengxiang Li and Tianxu Wang and Tengyu Liu and Xiaojian Ma and Yixin Chen and Baoxiong Jia and Siyuan Huang and Qing Li},
  year          = {2024},
  eprint        = {2408.04034},
  archivePrefix = {arXiv},
  primaryClass  = {cs.CV}
}

@InProceedings{Anderson_2018_CVPR,
  author    = {Anderson, Peter and Wu, Qi and Teney, Damien and Bruce, Jake and Johnson, Mark and S{\"u}nderhauf, Niko and Reid, Ian and Gould, Stephen and van den Hengel, Anton},
  title     = {Vision-and-Language Navigation: Interpreting Visually-Grounded Navigation Instructions in Real Environments},
  booktitle = {Proceedings of the IEEE Conference on Computer Vision and Pattern Recognition (CVPR)},
  month     = {June},
  year      = {2018},
  pages     = {3674--3683}
}

@InProceedings{Qi_2020_CVPR,
  author    = {Qi, Yuankai and Wu, Qi and Anderson, Peter and Wang, Xin and Wang, William Yang and Shen, Chunhua and Hengel, Anton van den},
  title     = {{REVERIE}: Remote Embodied Visual Referring Expression in Real Indoor Environments},
  booktitle = {Proceedings of the IEEE/CVF Conference on Computer Vision and Pattern Recognition (CVPR)},
  month     = {June},
  year      = {2020},
  pages     = {9982--9991}
}

@inproceedings{Krantz_2020_ECCV,
  author    = {Krantz, Jacob and Wijmans, Erik and Majumdar, Arjun and Batra, Dhruv and Lee, Stefan},
  title     = {Beyond the Nav-Graph: Vision-and-Language Navigation in Continuous Environments},
  booktitle = {Computer Vision -- ECCV 2020},
  series    = {Lecture Notes in Computer Science},
  volume    = {12373},
  pages     = {104--120},
  publisher = {Springer},
  year      = {2020},
  doi       = {10.1007/978-3-030-58604-1_7}
}

@misc{miao2026langmaphierarchicalbenchmarkopenvocabulary,
  title         = {{LangMap}: A Hierarchical Benchmark for Open-Vocabulary Goal Navigation},
  author        = {Bo Miao and Weijia Liu and Jun Luo and Lachlan Shinnick and Jian Liu and Thomas Hamilton-Smith and Yuhe Yang and Zijie Wu and Vanja Videnovic and Feras Dayoub and Anton van den Hengel},
  year          = {2026},
  eprint        = {2602.02220},
  archivePrefix = {arXiv},
  primaryClass  = {cs.CV}
}

@misc{chen2025searchingspacetimeunified,
  title         = {Searching in Space and Time: Unified Memory-Action Loops for Open-World Object Retrieval},
  author        = {Taijing Chen and Sateesh Kumar and Junhong Xu and Georgios Pavlakos and Joydeep Biswas and Roberto Mart{\'i}n-Mart{\'i}n},
  year          = {2025},
  eprint        = {2511.14004},
  archivePrefix = {arXiv},
  primaryClass  = {cs.RO}
}

@misc{dorbala2024personalizedembodiednavigation,
  title         = {Personalized Embodied Navigation for Portable Object Finding},
  author        = {Vishnu Sashank Dorbala and Bhrij Patel and Amrit Singh Bedi and Dinesh Manocha},
  year          = {2026},
  eprint        = {2403.09905},
  archivePrefix = {arXiv},
  primaryClass  = {cs.RO}
}

@misc{kolve2017ai2thor,
  title         = {{AI2-THOR}: An Interactive 3D Environment for Visual AI},
  author        = {Eric Kolve and Roozbeh Mottaghi and Winson Han and Eli VanderBilt and Luca Weihs and Alvaro Herrasti and Matt Deitke and Kiana Ehsani and Daniel Gordon and Yuke Zhu and Aniruddha Kembhavi and Abhinav Gupta and Ali Farhadi},
  year          = {2022},
  eprint        = {1712.05474},
  archivePrefix = {arXiv},
  primaryClass  = {cs.CV},
  doi           = {10.48550/arXiv.1712.05474}
}

@InProceedings{Cheng_2024_CVPR,
    author    = {Cheng, Tianheng and Song, Lin and Ge, Yixiao and Liu, Wenyu and Wang, Xinggang and Shan, Ying},
    title     = {YOLO-World: Real-Time Open-Vocabulary Object Detection},
    booktitle = {Proceedings of the IEEE/CVF Conference on Computer Vision and Pattern Recognition (CVPR)},
    month     = {June},
    year      = {2024},
    pages     = {16901-16911}
}

@InProceedings{Kirillov_2023_ICCV,
    author    = {Kirillov, Alexander and Mintun, Eric and Ravi, Nikhila and Mao, Hanzi and Rolland, Chloe and Gustafson, Laura and Xiao, Tete and Whitehead, Spencer and Berg, Alexander C. and Lo, Wan-Yen and Dollar, Piotr and Girshick, Ross},
    title     = {Segment Anything},
    booktitle = {Proceedings of the IEEE/CVF International Conference on Computer Vision (ICCV)},
    month     = {October},
    year      = {2023},
    pages     = {4015-4026}
}

@InProceedings{Cherti_2023_CVPR,
    author    = {Cherti, Mehdi and Beaumont, Romain and Wightman, Ross and Wortsman, Mitchell and Ilharco, Gabriel and Gordon, Cade and Schuhmann, Christoph and Schmidt, Ludwig and Jitsev, Jenia},
    title     = {Reproducible Scaling Laws for Contrastive Language-Image Learning},
    booktitle = {Proceedings of the IEEE/CVF Conference on Computer Vision and Pattern Recognition (CVPR)},
    month     = {June},
    year      = {2023},
    pages     = {2818-2829}
}

@inproceedings{pan2024langnav,
  title={Langnav: Language as a perceptual representation for navigation},
  author={Pan, Bowen and Panda, Rameswar and Jin, SouYoung and Feris, Rogerio and Oliva, Aude and Isola, Phillip and Kim, Yoon},
  booktitle={Findings of the Association for Computational Linguistics: NAACL 2024},
  pages={950--974},
  year={2024}
}

@inproceedings{
yang2025embodiedbench,
title={EmbodiedBench: Comprehensive Benchmarking Multi-modal Large Language Models for Vision-Driven Embodied Agents},
author={Rui Yang and Hanyang Chen and Junyu Zhang and Mark Zhao and Cheng Qian and Kangrui Wang and Qineng Wang and Teja Venkat Koripella and Marziyeh Movahedi and Manling Li and Heng Ji and Huan Zhang and Tong Zhang},
booktitle={Forty-second International Conference on Machine Learning},
year={2025},
url={https://openreview.net/forum?id=DgGF2LEBPS}
}

@inproceedings{gu2024conceptgraphs,
  title={Conceptgraphs: Open-vocabulary 3d scene graphs for perception and planning},
  author={Gu, Qiao and Kuwajerwala, Ali and Morin, Sacha and Jatavallabhula, Krishna Murthy and Sen, Bipasha and Agarwal, Aditya and Rivera, Corban and Paul, William and Ellis, Kirsty and Chellappa, Rama and others},
  booktitle={2024 IEEE International Conference on Robotics and Automation (ICRA)},
  pages={5021--5028},
  year={2024},
  organization={IEEE}
}

@misc{ren2024grounded,
      title={Grounded SAM: Assembling Open-World Models for Diverse Visual Tasks}, 
      author={Tianhe Ren and Shilong Liu and Ailing Zeng and Jing Lin and Kunchang Li and He Cao and Jiayu Chen and Xinyu Huang and Yukang Chen and Feng Yan and Zhaoyang Zeng and Hao Zhang and Feng Li and Jie Yang and Hongyang Li and Qing Jiang and Lei Zhang},
      year={2024},
      eprint={2401.14159},
      archivePrefix={arXiv},
      primaryClass={cs.CV}
}

@misc{matterport_maskrcnn_2017,
  title={Mask R-CNN for object detection and instance segmentation on Keras and TensorFlow},
  author={Waleed Abdulla},
  year={2017},
  publisher={Github},
  journal={GitHub repository},
  howpublished={\url{https://github.com/matterport/Mask_RCNN}},
}

@INPROCEEDINGS{1308004,
  author={Wolf, D. and Sukhatme, G.S.},
  booktitle={IEEE International Conference on Robotics and Automation, 2004. Proceedings. ICRA '04. 2004}, 
  title={Online simultaneous localization and mapping in dynamic environments}, 
  year={2004},
  volume={2},
  number={},
  pages={1301-1307 Vol.2},
  doi={10.1109/ROBOT.2004.1308004}}

@ARTICLE{9345478,
  author={Liu, Bo and Xiao, Xuesu and Stone, Peter},
  journal={IEEE Robotics and Automation Letters}, 
  title={A Lifelong Learning Approach to Mobile Robot Navigation}, 
  year={2021},
  volume={6},
  number={2},
  pages={1090-1096},
  doi={10.1109/LRA.2021.3056373}}

@inproceedings{shi2019openlorisscene,
    title={Are We Ready for Service Robots? The {OpenLORIS-Scene} Datasets for Lifelong {SLAM}},
    author={Xuesong Shi and Dongjiang Li and Pengpeng Zhao and Qinbin Tian and Yuxin Tian and Qiwei Long and Chunhao Zhu and Jingwei Song and Fei Qiao and Le Song and Yangquan Guo and Zhigang Wang and Yimin Zhang and Baoxing Qin and Wei Yang and Fangshi Wang and Rosa H. M. Chan and Qi She},
    booktitle={2020 International Conference on Robotics and Automation (ICRA)},
    year={2020},
    pages={3139-3145},
}

@article{Bescos_2018,
   title={DynaSLAM: Tracking, Mapping, and Inpainting in Dynamic Scenes},
   volume={3},
   ISSN={2377-3774},
   url={http://dx.doi.org/10.1109/LRA.2018.2860039},
   DOI={10.1109/lra.2018.2860039},
   number={4},
   journal={IEEE Robotics and Automation Letters},
   publisher={Institute of Electrical and Electronics Engineers (IEEE)},
   author={Bescos, Berta and Facil, Jose M. and Civera, Javier and Neira, Jose},
   year={2018},
   month=Oct, pages={4076–4083} }

@misc{qian2023povslamprobabilisticobjectawarevariational,
      title={POV-SLAM: Probabilistic Object-Aware Variational SLAM in Semi-Static Environments}, 
      author={Jingxing Qian and Veronica Chatrath and James Servos and Aaron Mavrinac and Wolfram Burgard and Steven L. Waslander and Angela P. Schoellig},
      year={2023},
      eprint={2307.00488},
      archivePrefix={arXiv},
      primaryClass={cs.RO},
      url={https://arxiv.org/abs/2307.00488}, 
}


\clearpage
\appendix
\section{Benchmark Details}
\label{app:benchmark-details}

\subsection{EvoNav-Bench Task Generation Pipeline}
\label{app:task-generation-pseudocode}

Algorithm~\ref{alg:appendix-generation-pseudocode} abstracts the stage-wise generation pipeline described in Sec.~\ref{subsec:task-generation}.
Here, $H$ is the source ProcTHOR~\cite{NEURIPS2022_27c546ab} house, $K$ is the number of stages, and $\Theta$ is the fixed generation hyperparameter set listed in Table~\ref{tab:generation-hyperparameters}.
The stage house $H^{(k)}$ is obtained by applying a scene-change record $\Delta H_k$ to the previous house state; \textsc{SampleSceneEvolution} corresponds to the room-selection and room-modification steps in Fig.~\ref{fig:evolve-pipeline}, and returns an empty $\Delta H_k$ when evolution is not triggered.
\textsc{SampleStageTask} corresponds to goal sampling: it returns the evaluator-side stage task $\tau_k$, including the target object and reference pose, together with the public goal input $g_k$, which contains text descriptions and the reference image.
The generation state $\mathcal{S}$ stores reference-trajectory-derived information used only during generation, including visited rooms, recent changed rooms, recent goal rooms, and the previous reference pose.
The output manifest $\mathcal{M}$ records the public stage input, evaluator-only oracle fields, and $\Delta H_k$, while $\mathcal{H}$ stores the generated per-stage house configurations.

\begin{algorithm}[H]
    \caption{\textbf{EvoNav-Bench task generation.}}
    \label{alg:appendix-generation-pseudocode}
    \small
    \begin{algorithmic}[1]
        \Require ProcTHOR house $H$, number of stages $K$, generation hyperparameters $\Theta$
        \Ensure Sequence manifest $\mathcal{M}$ and stage houses $\mathcal{H}$
        \State Initialize $H^{(0)} \gets H$, sequence manifest $\mathcal{M}$, stage-house set $\mathcal{H} \gets \emptyset$, and generation state $\mathcal{S}$
        \For{$k = 1$ to $K$}
            \If{$k=1$}
                \State $\Delta H_k \gets \emptyset$
            \Else
                \State $\Delta H_k \gets \Call{SampleSceneEvolution}{H^{(k-1)}, \mathcal{S}, \Theta}$
            \EndIf
            \State $H^{(k)} \gets \Call{Update}{H^{(k-1)}, \Delta H_k}$
            \State $s_k \gets \Call{SelectStartPose}{H^{(k)}, \mathcal{S}}$
            \State $(\tau_k, g_k) \gets \Call{SampleStageTask}{H^{(k)}, s_k, \mathcal{S}, \Theta}$
            \State $\mathcal{M} \gets \Call{AppendStageRecord}{\mathcal{M}, \Delta H_k, \tau_k, g_k, \mathcal{S}}$
            \State $\mathcal{H} \gets \mathcal{H} \cup \{H^{(k)}\}$
            \State $\mathcal{S} \gets \Call{UpdateGenerationState}{\mathcal{S}, H^{(k)}, s_k, \tau_k}$
        \EndFor
        \State \Return $(\mathcal{M}, \mathcal{H})$
    \end{algorithmic}
\end{algorithm}

\subsection{Generation Hyperparameters}
\label{app:generation-hyperparameters}

The generator uses the same fixed hyperparameter set $\Theta$ for all generated sequences.
Table~\ref{tab:generation-hyperparameters} lists each value and the generation step where it is applied.
These hyperparameters control stage evolution, room sampling, and goal sampling, and therefore affect the final generated data distribution.

\begin{table}[H]
    \centering
    \scriptsize
    \setlength{\tabcolsep}{3pt}
    \captionsetup{justification=raggedright,singlelinecheck=false}
    \caption{\textbf{Generation hyperparameters.} 
    Each hyperparameter is applied at the corresponding pipeline step described in Sec.~\ref{subsec:task-generation} and Appendix~\ref{app:task-generation-pseudocode}.}
    \label{tab:generation-hyperparameters}
    \begin{tabular}{@{}>{\raggedright\arraybackslash}p{0.28\linewidth}>{\raggedright\arraybackslash}p{0.46\linewidth}>{\raggedright\arraybackslash}p{0.14\linewidth}@{}}
        \toprule
        Pipeline step & Hyperparameter & Value \\
        \midrule
        Stage transition & $p^\text{evolve}$ & 0.5 \\
        \midrule
        Generation state & Changed-room history window & 2 \\
        Generation state & Goal-room history window & 2 \\
        \midrule
        Evolved-room sampling & $f^{\mathrm{visited}}$ & 3.0 \\
        Evolved-room sampling & $f^{\mathrm{evolved}}$ & 0.5 \\
        \midrule
        Evolution-type sampling & \texttt{SALIENT\_OBJECT} & 0.5 \\
        Evolution-type sampling & \texttt{RETAIN\_GROUP} & 0.4 \\
        Evolution-type sampling & \texttt{FULL\_SHUFFLE} & 0.1 \\
        \midrule
        Goal-room sampling & Changed-room multiplier & 9.0 \\
        Goal-room sampling & Previous-goal-room multiplier & 0.2 \\
        Goal-object sampling & Moved-object multiplier & 9.0 \\
        \bottomrule
    \end{tabular}
\end{table}

\subsection{Task Manifest}
\label{app:task-manifest}

The task manifest is organized as a sequence-level header followed by one record per stage.
The sequence-level header stores dataset identity, source-scene identity, random seed, number of stages, split, and the fixed generation parameters $\Theta$.
Each stage record corresponds to a stage house file, and separates the public stage input $g_k$ from evaluator-only metadata. 
$g_k$ contains the reference start pose $s_k$, text goal descriptions and the associated reference image.
The evaluator-only portion records the hidden scene-change record $\Delta H_k$, a snapshot of the generation state $\mathcal{S}$, and the oracle stage task $\tau_k$.
The oracle task includes the target instance, target category, target room, and the reference pose.
During evaluation, only $g_k$ is provided to the agent, while the rest of the stage record is hidden and used for computing metrics and diagnostics.

\subsection{Goal Description Generation}
\label{app:goal-description-generation}

For each sampled target object, the generator first selects a reference pose for constructing the visual goal input.
It enumerates reachable agent positions in the target room, keeps candidates whose horizontal distance to the target is 1.0--1.5 meters, and accepts a pose only when the target instance is visible from that pose.
If no such pose exists for the sampled target, the generator rejects the candidate and samples another target.
A reference RGB image is then rendered from the accepted pose at a resolution of 300$\times$300 pixels, using the ProcTHOR default embodiment.

Inspired by EmbodiedBench-style~\cite{yang2025embodiedbench} multimodal goal specification, we create three instructions for the same target given a reference image.
The \texttt{basic\_instruction} is generated deterministically from the ProcTHOR object category and relation context, for example ``Please navigate to the mug in the kitchen on the countertop next to the sink.''.
The relation context may include the room type when the category is not globally unique, the support object when the target is on a receptacle, and nearby anchors within fixed distance thresholds; the implementation keeps at most two nearby anchors, one structural anchor, and one wall-mounted anchor.
The \texttt{appearance\_instruction} and \texttt{unified\_instruction} are generated from the same reference image using \texttt{gpt-4o} with the following prompts.

\begin{center}
\setlength{\fboxsep}{5pt}
\fbox{%
\begin{minipage}{0.92\linewidth}
\small
\textbf{Appearance prompt.}
Given the target category, the basic instruction, and the reference image, write one short navigation instruction that identifies the target mainly by visible appearance, including shape and color first, and material, texture, size, or distinctive parts when helpful.
Do not mention coordinates, object IDs, simulator terms, or invisible attributes.
\end{minipage}}

\vspace{0.45em}
\fbox{%
\begin{minipage}{0.92\linewidth}
\small
\textbf{Unified prompt.}
Given the basic instruction, the appearance instruction, and the reference image, write one concise navigation instruction that combines spatial relations with visual appearance.
Keep relations such as \textit{on}, \textit{next to}, \textit{near}, or \textit{below} only when they are supported by the image.
\end{minipage}}
\end{center}

\subsection{Comparison with Related Navigation Benchmarks}
\label{app:ln-benchmark-comparison}

Table~\ref{tab:ln-benchmark-comparison} focuses on the lifelong navigation benchmarks and temporally dynamic benchmarks discussed in the main paper.
We compare them along six task properties: whether the setting is lifelong, whether agents navigate in continuous environments, whether the environment is unknown to the agent, whether the task includes temporal environment or object evolution, whether it changes the scene layout, and whether goals are specified through multiple modalities.
A \checkmark indicates that the corresponding property is part of the benchmark setting; $\triangle$ denotes local object-layout changes while preserving room topology and overall navigability.

\begin{table}[H]
    \centering
    \scriptsize
    \setlength{\tabcolsep}{2pt}
    \caption{\textbf{Diagnostic comparison with related benchmarks.}
    EvoNav-Bench combines lifelong navigation with scene evolution and local object-layout change.}
    \label{tab:ln-benchmark-comparison}
    \begin{tabular}{@{}>{\raggedright\arraybackslash}m{0.21\linewidth}*{6}{>{\centering\arraybackslash}m{0.115\linewidth}}@{}}
        \toprule
        Benchmark & Lifelong & Continuous & \shortstack{Unknown\\env.} & \shortstack{Temporal\\evolution} & \shortstack{Layout\\change} & \shortstack{Multimodal\\goal} \\
        \midrule
        MultiON~\cite{wani2020multion} & \checkmark & \checkmark & \checkmark & $\times$ & $\times$ & $\times$ \\
        GOAT-Bench~\cite{Khanna_2024_CVPR} & \checkmark & \checkmark & \checkmark & $\times$ & $\times$ & \checkmark \\
        C-Nav~\cite{NEURIPS2025_b78fef5a} & \checkmark & \checkmark & \checkmark & $\times$ & $\times$ & $\times$ \\
        LENL~\cite{wang2026lifelong} & \checkmark & \checkmark & \checkmark & $\times$ & $\times$ & $\times$ \\
        Portable ObjectNav~\cite{dorbala2024personalizedembodiednavigation} & $\times$ & $\times$ & $\times$ & \checkmark & $\times$ & $\times$ \\
        STARBench~\cite{chen2025searchingspacetimeunified} & $\times$ & \checkmark & $\times$ & \checkmark & $\times$ & $\times$ \\
        \midrule
        EvoNav-Bench & \checkmark & \checkmark & \checkmark & \checkmark & $\triangle$ & \checkmark \\
        \bottomrule
    \end{tabular}
\end{table}

\section{Evaluation and Implementation Details}
\label{app:evaluation-implementation}

\subsection{Multi-stage Evaluation Protocol}
\label{app:multi-stage-protocol}

Evaluation replays the stage houses and public stage inputs stored in the manifest, but uses the evaluated agent's own trajectory to connect stages.
At stage $k=1$, the agent is initialized at the manifest start pose.
For every later stage, the evaluator first records the terminal pose from stage $k-1$, loads the next stage house $H^{(k)}$, and initializes the agent at that recorded pose.
If the recorded position is no longer reachable after the hidden scene change, the evaluator snaps it to the nearest reachable position while preserving the carried-over orientation.
This fallback is used only to keep the physical rollout valid after local object relocation; the requested and effective start poses are both recorded for diagnostics.

At the beginning of each stage, the agent receives only the public goal input $g_k$.
In the reported experiments, the stage input is the \texttt{unified\_instruction}.
The scene-change record $\Delta H_k$, target instance, reference pose, and other oracle fields remain evaluator-only.
The stage terminates when the agent stops or when the stage budget is exhausted, and the next stage is released regardless of whether the stop is evaluated as successful.
The budget is reset at each stage boundary, while the agent's internal state is preserved unless a baseline variant explicitly resets part of it.
No observation is provided during the stage transition itself, so an agent can discover scene evolution only through subsequent RGB-D observations in the new stage.
This evaluation transition is separate from the reference trajectory used during task generation: the reference trajectory fixes the stage schedule, whereas evaluation uses the agent's actual movement.

\subsection{Metric Computation}
\label{app:metric-computation}

We compute all reported metrics at the stage level.
Let $\mathcal{K}$ be the set of evaluated stages, $p_k^{\mathrm{term}}$ be the agent pose at the end of stage $k$, and $x_k$ be the target.
Stage success is defined by the evaluator-side distance condition at termination:
\begin{equation}
    S_k =
    \mathbb{I}\left[
        d_{\mathrm{2D}}\left(p_k^{\mathrm{term}}, x_k\right)
        < r_{\mathrm{succ}}
    \right],
\end{equation}
where $d_{\mathrm{2D}}$ is Euclidean distance on the simulator $x$--$z$ ground plane, and $r_{\mathrm{succ}}=2.0\,\mathrm{m}$ is the success radius.

Let $L_k$ denote the executed path length in stage $k$, and let $L_k^\star$ denote the evaluator reference path length from the effective stage start pose to $x_k$ in the corresponding stage house.
The evaluator queries AI2-THOR~\cite{kolve2017ai2thor} for a shortest path to $x_k$.
The per-stage SPL is
\begin{equation}
    \mathrm{SPL}_k =
    S_k \frac{L_k^\star}{\max\left(L_k, L_k^\star\right)}.
\end{equation}
Finally, SR and SPL are averaged over stages:
\begin{equation}
    \mathrm{SR} =
    \frac{1}{|\mathcal{K}|}\sum_{k\in\mathcal{K}} S_k,
    \qquad
    \mathrm{SPL} =
    \frac{1}{|\mathcal{K}|}\sum_{k\in\mathcal{K}} \mathrm{SPL}_k.
\end{equation}

\subsection{Adapting Existing Agents to EvoNav-Bench}
\label{app:baseline-adaptation}

\textbf{UniGoal.}
We adapt UniGoal~\cite{Yin_2025_CVPR} by replacing its original simulator interface with the EvoNav-Bench stage protocol while preserving its graph-based navigation loop.
The agent uses the discrete action interface, with \texttt{MoveAhead} by 0.25\,m, \texttt{RotateLeft} by $15^\circ$, \texttt{RotateRight} by $15^\circ$, and \texttt{Stop} as its primitive actions.
We set the per-stage budget to 500 primitive actions.
Each stage maps the public goal through UniGoal's original goal-graph construction procedure into a target subgraph.
RGB-D observations are accumulated into the existing representations, but stale scene-graph entries and outdated occupied regions may also be retained.
We implement \texttt{gpt-4o} as UniGoal's VLM.
Under GT Perception, ProcTHOR object types are mapped into UniGoal's semantic-map and scene-graph category spaces.
Predicted Perception follows UniGoal's original Mask R-CNN~\cite{matterport_maskrcnn_2017} and Grounded-SAM~\cite{ren2024grounded} perception stack.

\textbf{3D-Mem.}
We retain 3D-Mem's~\cite{Yang_2025_CVPR} memory and decision-loop components, replacing the observation source with ProcTHOR RGB-D observations, camera poses, and detections.
We also replace the TSDF-based occupancy map with occupancy map extracted through ray-casting on oracle occupancy map provided by simulator. The original 3D-Mem implementation extracts occupancy map from a single horizontal slice of the TSDF map, which does not capture the agent’s full collision volume.
For example, a slice through the empty space beneath a tabletop may label the area as free even though the agent cannot traverse it.
The accumulated occupancy map and snapshot memory are not proactively reconstructed at stage boundaries, allowing both useful prior observations and stale evidence to persist.
3D-Mem's decision-level control structure is preserved: each decision round asks the VLM \texttt{gpt-4o} to select a snapshot or frontier, and the selected candidate is converted into a target position.
The decision-round budget is set to 50 per stage.
Because 3D-Mem does not directly output primitive actions, the adapter executes each selected target by following planner waypoints with AI2-THOR \texttt{TeleportFull} updates, preserving the original decision-round semantics.
GT Perception uses evaluator-provided object masks and semantic labels, while Predicted Perception follows the original 3D-Mem detector stack with YOLOv8x-World~\cite{Cheng_2024_CVPR}, SAM ViT-L~\cite{Kirillov_2023_ICCV}, and OpenCLIP~\cite{Cherti_2023_CVPR}.

\textbf{MSGNav.}
Since MSGNav~\cite{huang2026msgnavunleashingpowermultimodal} itself is built on 3D-Mem, we adapt it to EvoNav-Bench by following the same procedure as 3D-Mem. 
Following MSGNav, we also replace the text-only edge with image edges. We maintain the original MSGNav decision loop, and incorporate the four navigation modules: KSS, AVU, CLR and VVD.

\subsection{Heuristic Implementations}
\label{app:fu-variants}

\textbf{Frontier-Update (FU) Implementation.}
Frontier-Update (FU) follows the same high-level decision loop as 3D-Mem: 
it maintains an occupancy map, derives exploration frontiers, accumulates object entries, and asks the VLM to select navigation targets from object or frontier candidates.
FU adapts this loop to evolving scenes through two changes.
First, while following 3D-Mem's object creation procedure, it exposes object-level memory rather than snapshots as the primary evidence for object candidates, because a snapshot can bind an object to stale surrounding context after scene evolution.
Second, FU rebuilds the occupancy map and resets the frontier set when current observation contradicts the current occupancy map, enabling the agent to re-explore changed rooms while retaining reusable object entries.

Concretely, FU follows ConceptGraph~\cite{gu2024conceptgraphs} to perform a series of object segmentation, spatial transformations, matching, and merging, to produce an object entries with semantic labels, visual evidence, 3D locations, and point clouds.
FU also utilizes the occupancy map provided by the simulator. 
When incoming occupancy map conflicts with the accumulated occupancy map, FU treats the accumulated occupancy map as outdated, abandons the active navigation target, and regenerates frontiers from subsequent observations.
The next decision is then made over the retained object entries and the updated frontier set, allowing the agent to navigate to reliable object entries when available and to re-explore changed regions otherwise.
To keep the VLM input compact, FU uses CLIP similarity to pre-rank stored objects and forwards only the top five object entries to the VLM.
If the VLM selects an object whose location is not supported by the currently explored map, FU routes the agent to the nearest current frontier.
FU uses the same decision-round budget, waypoint-based \texttt{TeleportFull} execution, and GT/Predicted Perception settings as 3D-Mem.

\textbf{Stage-Reset Implementation.}
Stage-Reset shares a similar pipeline with FU.
However, at the beginning of each stage, Stage-Reset clears the accumulated object entries and occupancy map, forcing the agent to explore each stage from scratch.
Consequently, Stage-Reset does not retain any prior information across stages. Stage-Reset would neither benefit from reusable object entries nor suffer from stale entries after scene evolution, and therefore serves as a baseline for evaluating the effect of retained information.

\textbf{Fail-then-Update(FTU) Implementation.}
FTU is built upon FU.
FTU adds a verification step after the agent reaches the predicted location of a selected object entry.
If the VLM judges that the current observation does not support the existence of store object entry, FTU removes the entry and starts a new decision round.

\begin{figure}
    \centering
    \includegraphics[width=0.8\textwidth]{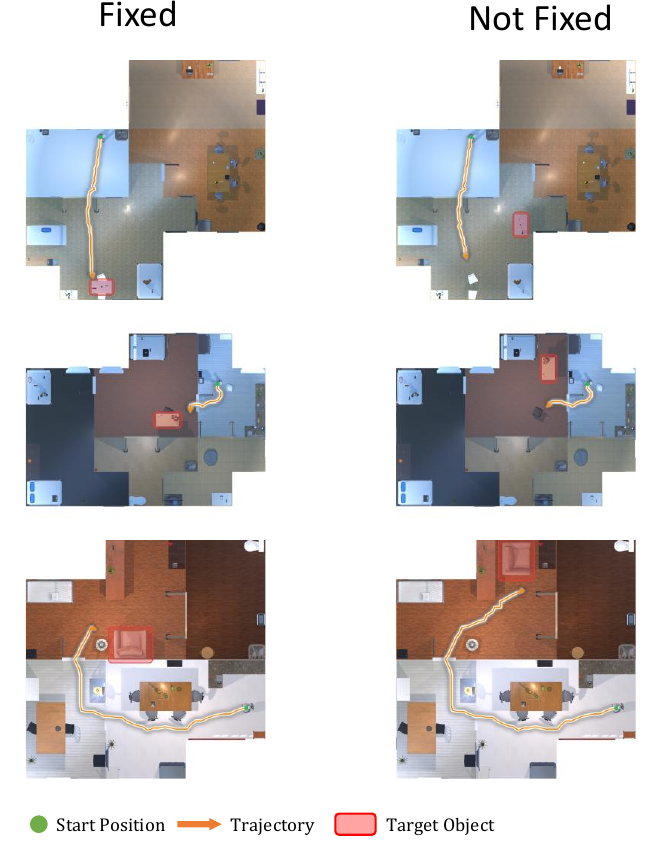}
    \caption{\textbf{Qualitative FU rollouts in the Fixed/Not Fixed controlled analysis.}
    Each row compares the Fixed condition (left) and the Not Fixed condition (right) for the same Stage-2 task.
    The two evaluations share the same Stage-1 reference trajectory and the same Stage-2 goal; following Sec.~\ref{sec:controlled-analysis}, this goal is sampled from objects observed along the shared reference trajectory.}
    \label{fig:compare-evolution-visualization}
    \vspace{-1.5em}
\end{figure}

\section{Visualizing the Effect of Environment Evolution}
\label{app:evolution-visualization}
To complement the controlled Fixed/Not Fixed analysis in Sec.~\ref{sec:controlled-analysis}, Fig.~\ref{fig:compare-evolution-visualization} provides qualitative FU rollouts under the same paired setting.
Each row compares the same Stage-2 task under two conditions: \textbf{Fixed}, where the scene remains unchanged after the Stage-1 reference trajectory, and \textbf{Not Fixed}, where the scene evolves before Stage-2 evaluation.
Because the two rollouts in each row share the same Stage-1 reference trajectory and the same Stage-2 goal, their trajectory differences mainly reflect whether scene evolution invalidates retained object entries.

In the Fixed condition, the target location remains consistent with the previous-stage observations, so retained object entries can guide FU directly toward the target.
In the Not Fixed condition, however, a moved target can leave a plausible but outdated object entry at its previous location.
The first two examples show this failure mode: FU follows stale retained evidence toward the old target location while the current target has moved elsewhere.
The last example shows a recovery case, where FU eventually reaches the updated target location through additional exploration.
Together, these examples visualize the mechanism behind the quantitative finding in Sec.~\ref{sec:controlled-analysis}: retained object entries are useful in fixed scenes but can become harmful after scene evolution.

\end{document}